\pdfoutput=1
\documentclass[11pt]{article}
\usepackage[final]{acl}
\usepackage{times}
\usepackage{latexsym}
\usepackage{listings}
\usepackage[utf8]{inputenc}
\usepackage{microtype}
\usepackage{inconsolata}

\usepackage{graphicx}
\usepackage{multirow}
\usepackage{algorithm}
\usepackage{algorithmic}
\usepackage{booktabs}
\usepackage{lipsum}
\usepackage[T2A,T1]{fontenc} 
\usepackage{textcomp}
\usepackage{newtxtt}

\makeatletter
\renewcommand{\verbatim@font}{\usefont{T2A}{lh}{m}{n}\small\setlength{\lineskip}{3pt}}
\makeatother

\usepackage{fancyvrb}

\usepackage{tgcursor}
\usepackage{alltt}

\title{Presence or Absence: \\
Are Unknown Word Usages in Dictionaries?}

\author{Xianghe Ma$^{1}$ \, Dominik Schlechtweg$^{2}$ \, 
Wei Zhao$^{3}$ \\[0.4em]
    $^{1}$University of Heidelberg\,  
    $^2$University of Stuttgart\,  
    $^3$University of Aberdeen \\
    \texttt{xianghe.ma@stud.uni-heidelberg.de}\\
    \texttt{dominik.schlechtweg@ims.uni-stuttgart.de}\\
    \texttt{wei.zhao@abdn.ac.uk}\\
  }

\begin{document}
\maketitle
\begin{abstract}
There has been a surge of interest in computational modeling of semantic change. The foci of previous works are on detecting and interpreting word senses gained over time; however, it remains unclear whether the gained senses are covered by dictionaries. In this work, we aim to fill this research gap by comparing detected word senses with dictionary sense inventories in order to bridge between the communities of lexical semantic change detection and lexicography. We evaluate our system in the AXOLOTL-24 shared task for Finnish, Russian and German languages \cite{fedorova-etal-2024-axolotl}. Our system is fully unsupervised. It leverages a graph-based clustering approach to predict mappings between unknown word usages and dictionary entries for Subtask 1, and generates dictionary-like definitions for those novel word usages through the state-of-the-art Large Language Models such as GPT-4 and LLaMA-3 for Subtask 2. In Subtask 1, our system outperforms the baseline system by a large margin, and it offers interpretability for the mapping results by distinguishing between matched and unmatched (novel) word usages through our graph-based clustering approach. Our system ranks first in Finnish and German, and ranks second in Russian on the Subtask 2 test-phase leaderboard. These results show the potential of our system in managing dictionary entries, particularly for updating dictionaries to include novel sense entries. Our code and data are made publicly available\footnote{\url{https://github.com/xiaohemaikoo/axolotl24-ABDN-NLP}}.
\end{abstract}

\section{Introduction}

Meaning changes over time have been a subject of research for many years in historical linguistics \citep[e.g.][]{Blank97XVI,Geeraerts2020Smurf}.
Researchers use linguistic tools and methods to identify gained and lost meanings of headwords, and more importantly to interpret these changes by categorizing the types of changes and detecting social and cultural forces driving the changes. 

\begin{figure}[htb]
\centering
\includegraphics[width=\linewidth]{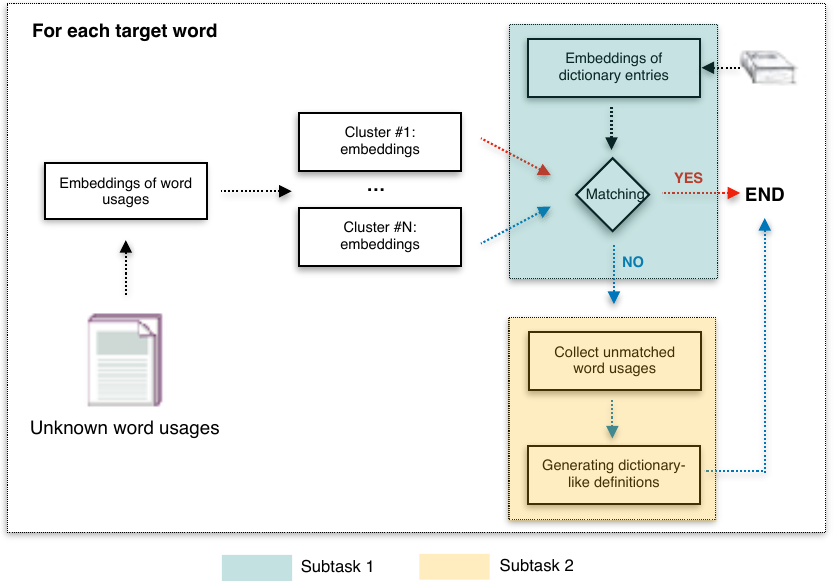}
\caption{An illustration of the workflow for the two AXOLOTL-24 subtasks. Unknown word usages refer to usages found at a later time period, and their mappings with dictionary sense entries are unknown.}
\label{fig:workflow}
\end{figure}

Recently, there has been scholarly interest in computational modeling of meaning changes as  cost-efficient alternatives to labor-intensive linguistic tools and methods. As a result, a plateau of research outputs has been made, including shared tasks and datasets \citep[e.g.][]{schlechtweg2020semeval, kutuzov2021rushifteval, d-zamora-reina-etal-2022-black, chen-etal-2023-chiwug-graph, schlechtweg2024lscd}, models \cite{eger-mehler-2016-linearity, hamilton-etal-2016-cultural, hamilton-etal-2016-diachronic, 
martinc2020capturing, kaiser2021effects, 
montariol-etal-2021-scalable,teodorescu2022ualberta, cassotti-etal-2023-xl, ma-etal-2024-graph}, tools \citep[][]{schlechtweg2024dureltool}, and relevant workshops\footnote{\url{https://www.changeiskey.org/event/2024-acl-lchange/}}. For instance,  SemEval2020 Task 1 \cite{schlechtweg2020semeval}, a seminal work on this topic, introduces the first task and datasets on unsupervised lexical semantic change detection in English, German, Swedish and Latin languages. Further extensions include DIACR-Ita for Italian \citep[][]{diacrita_evalita2020}, RuShiftEval for Russian \cite{kutuzov2021rushifteval}, and LSCDiscovery for Spanish \cite{d-zamora-reina-etal-2022-black}.

The immediate impact of these research outputs might be on the lexicography industry. Lexicographers rely on collocations and grammatical patterns to identify novel meanings that are not included in dictionaries, and add these identified meanings into the next iteration of dictionary updates \cite{kilgarriff2010semi}. However, the process of doing so is costly and time-consuming. For instance, in 2023, the Oxford English Dictionary created about 1,700 new meanings\footnote{\url{https://www.oed.com/information/updates}}, with the help of hundreds of language specialists for English alone. Recently, the AXOLOTL-24 shared task has connected lexical semantic change detection with dictionary entries. Instead of just detecting meaning change, the shared task aims to align dictionary sense entries with each word usage. This is particularly useful for managing dictionary entries, e.g., to identify and collect novel meanings not covered by dictionaries \citep[][]{erk-2006-unknown,Lautenschlager2024nonrecorded}. 	

In this work, we participate in two AXOLOTL-24 subtasks for Finnish, Russian and German languages. The tasks include (a) bridging diachronic word uses and a synchronic dictionary and (b) definition generation for novel word senses. The first subtask aims to predict mappings between dictionary meaning entries and word usages while the second task plans to produce dictionary-like definitions for those unmatched usages with novel word meanings not covered by dictionaries. In the following, we outline the components of our system:
\begin{itemize}
    \item For Subtask 1, we keep the workflow of the AXOLOTL-24 baseline system unchanged, which includes three components: producing embeddings for word usages, clustering these embeddings, and mapping between dictionary meaning entries and the resulting clusters. However, we make modifications to each component. The component-wise system comparison is presented in Table \ref{tab:comparison-task1}.

    \item For Subtask 2, unlike the baseline system, which requires costly model training for generating dictionary-like definitions for unmatched word usages, our system is training-free and does so by just prompting Large Language Models such as GPT-4 \citep{achiam2023gpt} and LLaMA-3\footnote{\url{https://llama.meta.com/llama3/}}. We provide the system comparison in Table \ref{tab:comparison-task2}.
    
\end{itemize}

\section{Related Work}
This section reviews semantic change detection and discuss its potential connections with dictionaries.
\paragraph{Lexical Semantic Change Detection (LSCD)} focuses on the automatic identification of shifts in word meanings over time. For instance, the word `chill' used to mean `cold' for individuals growing up in the 60s, but for those in the 90s, it means `relaxed'. Many works proposed to detect meaning shifts by using static or contextualized embeddings 
\citep[][]{eger-mehler-2016-linearity, hamilton-etal-2016-cultural, hamilton-etal-2016-diachronic, 
martinc2020capturing, gonen-etal-2020-simple, kaiser2021effects, 
montariol-etal-2021-scalable,teodorescu2022ualberta,deepmistake-lcsdiscovery}. Most work in LSCD has been done on an unsupervised task formulation \citep[][]{schlechtweg2020semeval} which neither involved a dictionary, nor providing interpretation or qualification of detected sense changes. While early work on static embeddings \citep[][]{Kim14,hamilton2016diachronic} could qualify changes to a certain extent through nearest neighbors, it usually did not provide sense clusters in a dictionary-like manner. More recent work straightforwardly enables the induction of sense clusters through clustering of contextualized embeddings \citep[][]{giulianelli-etal-2020-analysing,bos-lcsdiscovery,Montariol2021scalable,Arefyev2021Interpretable}. More recently, \citet{ma-etal-2024-graph} presented a graph-based clustering approach to detect gained word senses with low frequency, and offered 
interpretability by visualizing cross-language semantic changes. 
The works by \citet{giulianelli-etal-2023-interpretable,Kutuzov2024enriching}
offer new ways of interpretability such as automatically generating sense definitions for usages from clusters.
For an overview of recent model architectures incl. clustering approaches, see \citet[][]{d-zamora-reina-etal-2022-black}.

\paragraph{LSCD and dictionaries.} The above-described approaches all have in common that they do not involve a dictionary in their task formulation. However, a variety of dictionaries is available for different languages and time periods \citep[e.g.][]{dal1955explanatory,Paul02XXI,oxford2009oxford} providing valuable information characterizing a language stage on the lexical level. Thus, a possible alternative task formulation for LSCD is to start from an existing dictionary and compare corpus usages against the dictionary entries in order to find usages not covered by the dictionary \citep[][]{erk-2006-unknown,Lautenschlager2024nonrecorded}.

\section{AXOLOTL-24 Shared Task}
Participants are asked to solve the two subtasks:
\begin{itemize}
    \item \textbf{Subtask 1 - bridging diachronic word
uses and a synchronic dictionary}: 
This task is to identify mappings between dictionary entries and the word usages of each target word, i.e., that the task asks to detect whether each word usage has a novel sense or not, meaning that it is not (or is) recorded in dictionaries.

    \item \textbf{Subtask 2 - definition generation for novel word senses}: 
    This task builds upon the mapping results of Subtask 1. It aims to generate dictionary-like definitions for the unmatched word usages discovered in Subtask 1, i.e., that these usages contain novel senses not covered by dictionaries.

\end{itemize}
An example for the Finnish target word `palaus' is illustrated in Figure \ref{fig:running-example}. Participants are provided with the mappings of usages at an earlier time period to dictionary entries (sense glosses) while the mappings for a later time period is unknown. Subtask 1 asks participants to predict which sense gloss Usage 3 belongs to. If a system predicts Usage 3 to have a novel sense not covered by existing sense glosses, then Subtask 2 asks to generate the gloss for the novel sense.

\begin{figure}[h]   
    \begin{minipage}{\textwidth}
    \begin{verbatim}
[Gloss 1]: kääntymys, hengellinen kääntyminen

[Gloss 2]: kuumuus

[Word Usage 1] (<1700): anna minulle yxi oikea 
catumus ia synnistä palaus. 

[Word Usage 2] (<1700): Coska nyt Pauali cocosi 
ydhen coghon Risuija ia pani ne Tulen päle, edesmateli 
yxi Kyykerme palaudhesta.

[Word Usage 3] (>1700): Jumala on itze joca meisä sen 
suuren Palauxen ja muutoxen toimitta

[Mapping]: (Usage 1, Gloss 1), 
               (Usage 2, Gloss 2),
               (Usage 3, [Gloss 1, Gloss 2, Unknown])
    \end{verbatim}    
    \end{minipage}
    \caption{A running example for the target word `palaus' from the Finnish test set. The first two usages (before 1700) belong to the earlier time period while the last one belongs to the later.}
    \label{fig:running-example}
\end{figure}

\section{Our Systems}
\subsection{Subtask 1}

\paragraph{Workflow.} We reuse the workflow of the AXOLOTL-24 baseline system, which includes the following three components that are executed sequentially:
\begin{itemize}
    \item \textbf{Producing embeddings of word usages}: This component aims to encode the usages of a target word.
    
    \item \textbf{Clustering embeddings}: This component is to partition the resulting embeddings of a target word into clusters. Each cluster contains embeddings with similar meanings.
   
    \item \textbf{Mapping between dictionary sense entries and clusters}: This component is to align dictionary sense entries with the resulting clusters. If the semantic meaning represented by a cluster is present in dictionaries, then we assign the dictionary entry to that cluster. Otherwise, a novel meaning is said to be identified. This implies the need for dictionary updates to include new sense entries.
\end{itemize}

\paragraph{Baseline.} The baseline system proposes an unsupervised approach that does not rely on training data, i.e., the lack of mappings between word usages at an earlier time period and dictionary sense entries, to predict mappings for unknown word usages at a later period. The idea for the baseline system to implement the workflow is the following: 
For each target word, the baseline system begins with collecting all the relevant 
corpus usages
available at an earlier time period. If corpus usages are unavailable\footnote{For the Russian datasets in the AXOLOTL-24 shared task, some corpus usages in the 19th century are missing.}, the system resorts to using dictionary definitions of the target word as substitutes. Secondly, the system aims to encode the meanings of the target word in various corpus usages. However, doing so is not trivial, as the positions of the target word in corpus usages are not always given in the AXOLOTL-24 datasets. Moreover, for morphologically rich languages, the automatic process of locating the target word in word usages is inaccurate.
Thus, the baseline system approaches the meaning of a target word by using the sentence encoder LEALLA \cite{mao-nakagawa-2023-lealla} to produce the embedding for the entire word usage.

After collecting word usage embeddings, 
the baseline system leverages a popular clustering approach known as Affinity Propagation \citep{frey2007clustering} to group word usage embeddings into several clusters. Each cluster contains multiple embeddings with similar meanings.

Lastly, to map between dictionary sense entries with unknown usages of a target word at a later time period, the baseline system proposes to align dictionary entries with the collective meaning of each cluster. In particular, for each cluster, the system chooses the embedding of the first-indexed usage of the target word in the AXOLOTL-24 datasets
as the collective meaning represented by that cluster. It then computes the cosine similarity between that word usage embedding and the embedding of each dictionary entry (i.e., sense gloss). If the similarity score surpasses a predefined threshold, then all the word usages within that cluster are said to be matching that dictionary entry. 

\begin{table}
\small
\centering
\setlength\tabcolsep{1.5pt}
\begin{tabular}{@{}lll@{}}
\toprule
\textbf{Components} & \textbf{Baseline} & \textbf{Our System} \\
\midrule
Embedding &  word usages & word usages and words \\
\midrule
Clustering & Affinity Prop. & Neighbor-based clustering \\
\midrule
Mapping & first-indexed emb. & average emb. \\
\bottomrule
\end{tabular}
\caption{Component-wise comparison between the baseline and our system in Subtask 1.}
\label{tab:comparison-task1}
\end{table}

\paragraph{Our submitted system.} 
Just like the baseline system, our system also does not rely on training data to predict mappings between unknown usages at a later time period and dictionary entries. However, we make substantial changes to each component of the workflow. 

\begin{figure}[h]
  \centering
  \includegraphics[width=0.5\textwidth]{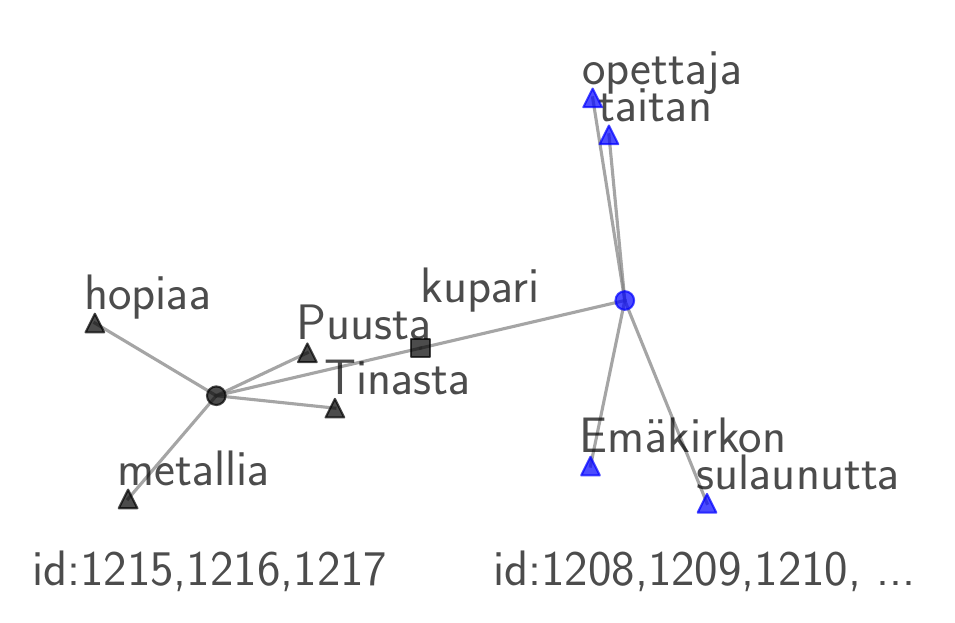}
  \caption{
  An illustration of our semantic graph for the Finnish target word `kupari' (root node in the graph), together with two subtrees separating two meaning clusters. One cluster represents the meaning related to a metal (in black) that is covered by dictionaries while the other represents the novel meaning `the recipient of metals as currency' (in blue) that is not. Each cluster contains 4-nearest neighboring words, together with their corpus usage IDs,  to interpret the collective meaning of the cluster.}  
  \label{fig:semantic-graph}
\end{figure}

For each target word, we produce word usage embeddings\footnote{For our system, a word usage embedding is defined as the average of all m-BERT word embeddings in a corpus usage.} by using m-BERT \cite{devlin2018bert} to encode various corpus usages of the target word. Moreover, we create a vocabulary containing all the words available in the entire corpus, together with their average BERT-based word embeddings over their occurrences in the corpus. We take all the word usage embeddings of a target word and the vocabulary as input to derive a 3-layer semantic graph for each target word through our clustering method. Each semantic graph contains the following elements:
\begin{itemize}
    \item \textbf{Root node} represents the average word usage embedding over all the usages of a target word in the corpus.
    \item \textbf{Nodes on the second layer} are centroids of each sense cluster, i.e., the average of word usage embeddings within each cluster. 
    \item \textbf{Nodes on the third layer} are k-nearest neighbors to each cluster centroid.
\end{itemize}
Note that our clustering only operates on embeddings, and the nodes on the second layer are built upon the clustering result. We introduce such a graph as a visualization tool after clustering to separate sense clusters and the corresponding word usages. See a two-dimensional illustration in Figure \ref{fig:semantic-graph}---where the graph separates a recorded word sense from an unrecorded (novel) sense, together with their word usages from the Finnish AXOLOTL-24 dev set.

Lastly, to map dictionary entries to clusters, our system differs from the baseline: Instead of choosing the first-indexed word usage embedding as the collective meaning of a cluster, our system does so by using the average word usage embedding. Here, we briefly outline our clustering approach. For further details, we refer to \citet{ma-etal-2024-graph}. 
\paragraph{Clustering.} 
For each target word $w$, we denote $\mathcal{C}_w = \{c_1, c_2, \dots, c_n\}$ as a word cloud consisting of a set of $d$-dimensional embeddings. Each embedding represents a corpus usage of the target word, and $n$ denotes the number of word usages available in a given corpus that contain that target word. We aim to partition $\mathcal{C}_w$ into $m$ clusters. Each cluster contains a subset of $\mathcal{C}_w$ representing embeddings of word usages with similar meanings. Our clustering method is illustrated in Algorithm \ref{alg:clustering}. We choose our clustering over the baseline Affinity Propagation \cite{frey2007clustering} because target words in the AXOLOTL-24 datasets have 2-23 usages on average (c.f. Table \ref{tab:AXOLOTL-24-dataset-statistics}), i.e., they only have low-frequency senses; in such a setup, our clustering largely outperforms Affinity Propagation (see Table 11 in \citet{ma-etal-2024-graph}). We present our clustering details in the following:

\begin{algorithm}[ht]
\centering
\caption{Our clustering method}\label{alg:clustering}
\begin{algorithmic}[1]
\REQUIRE $\mathcal{C}_w=\{c_i\}_{i=1}^n$ as a set of word usage embeddings representing various usages of a target word $w$, 
      $t_{sc}$ as the maximum distance between similar clusters.    
\STATE Initial centroids of clusters: $\mathcal{P}_w=\{p_i|p_i = c_i\}_{i=1}^n$
\WHILE{$\min_{p_i\in \mathcal{P}_w, p_j\in \mathcal{P}_w, i \neq j}d(p_i, p_j)  < t_{sc}$} 
\STATE $\mathcal{P}_w = (\mathcal{P}_w \setminus \{p_i, p_j\}) \cup \{\frac{p_i+p_j}{2}\}$ 
\ENDWHILE
\RETURN $\mathcal{P}_w$
\end{algorithmic}
\end{algorithm}

Our clustering method is similar to the bottom-up agglomerative clustering \cite{sibson1973slink} but differs in that we use a neighbor-based metric\footnote{For agglomerative clustering, the distance between two clusters is calculated as the average pairwise distance between usage pairs based on their embeddings. For us, each pairwise distance is calculated as the bipartite matching score over $k$-nearest neighbors of a word usage and those of another.} to handle low-frequency clusters. The idea is the following:
We start by treating each embedding
as a separate cluster, and then iteratively merge two clusters when their centroids are of a distance smaller than the distance threshold $t_{sc}$ until no further pairs of such similar clusters can be found. Following \citet{ma-etal-2024-graph}, we use a neighbor-based distance metric in the clustering process to compute distances between clusters. 
Both the distance threshold and the number of nearest neighbors are hyperparameters,
which we tune on dev sets.

Importantly, using a neighbor-based distance metric in the clustering process is crucial for handling many low-frequency word senses in the AXOLOTL-24 datasets. \citet{ma-etal-2024-graph} showed that using such a metric to compute distances between clusters is a contributing factor to identify low-frequency sense clusters. The reason for this is the following: for a low-frequency sense with few word usages, relying on those usages to decide whether they should form a standalone low-frequency cluster or be merged into another cluster can be unreliable. However, with k-nearest neighbors of those usages participating (i.e., additional information provided) in the decision making, the decision becomes more reliable.

Lastly, for mapping, we select the average usage embedding (i.e., cluster centroid) as the collective meaning of a cluster, and compare that embedding with dictionary entries. We choose the average embedding over the embedding of the first-indexed usage of a target word (see Table \ref{tab:comparison-task1}) because the first-indexed choice is almost random. We use the average embedding to eliminate such randomness.

\subsection{Subtask 2}

\paragraph{Workflow.} Our submitted system follows the workflow of the AXOLOTL-24 baseline that includes the two sequential components below:

\begin{itemize}
    \item \textbf{Collecting unmatched word usages.} This component aims to collect word usages with novel senses not found in dictionaries. Doing so is straightforward: The mapping results from Subtask 1 include word usages that match dictionary entries, as well as  unmatched (novel) usages. Here, we only collect those unmatched  usages.     
    We note that the system performance in Subtask 1 immediately impacts the quality of this component.
    \item \textbf{Generating definitions.} This component takes unmatched word usages as input and generates their dictionary-like definitions. 
\end{itemize}

\begin{table}
\small
\centering
\begin{tabular}{lll}
\toprule
\textbf{Components} & \textbf{Baseline} & \textbf{Our System} \\
\midrule
Collection & \multicolumn{2}{c}{collect the mapping results of Subtask 1}   \\
\midrule
Generation & finetune XGLM & prompte LLMs \\
\bottomrule
\end{tabular}
\caption{Component-wise comparison between the baseline and our system in Subtask 2.}
\label{tab:comparison-task2}
\end{table}

\paragraph{Baseline.} The baseline system proposes a supervised approach that trains a generative model on train sets, i.e., the mappings between dictionary entries and matched word usages, in order to generate definitions for unmatched word usages. 
In particular, the system takes a target word and its matched word usages as input, and dictionary definitions of these word usages as the ground-truth output. The system uses the generative model XGLM \cite{lin-etal-2022-shot} to encode the input and fine-tunes its model parameters by minimizing the cross-entropy loss in a way to make the generated definitions as close as possible to the ground-truth counterparts. Note that the fine-tuning process of the baseline is costly as it is executed separately for each language.  

\paragraph{Our submitted system.} Unlike the baseline system, our system is fully unsupervised\footnote{Our system based on LLMs is unsupervised in that it does not rely on training data; however, the training data for pre-training LLMs include many human-annotated data.}. After collecting unmatched word usages we prompt Large Language Models (LLMs) to generate definitions for these word usages. We experiment with several LLMs including open-source and commercial models (LLaMA and GPT). Figure \ref{fig:prompt} (appendix) illustrates the prompt to instruct GPT-3.5-turbo\footnote{\url{https://platform.openai.com/docs/models/gpt-3-5-turbo}} to generate English definitions. 

\section{Experiments}

\paragraph{Datasets.}

\begin{table*}[h]
\centering 
\footnotesize
\setlength\tabcolsep{3pt} 
\begin{tabular}{lccccccccccr}
\toprule
\multicolumn{1}{l}{} & \multicolumn{5}{c}{Corpus \#1}                & \multicolumn{5}{c}{Corpus \#2}                    & \multicolumn{1}{l}{} \\
 \cmidrule(lr){2-6}\cmidrule(lr){7-11}
Languages             &Period ($t-1$)    & \#usages & avg/u & max/u & min/u & Period ($t$)        & \#usages & avg/u & max/u & min/u & \#targets              \\
 \cmidrule(lr){1-1} \cmidrule(lr){2-6}\cmidrule(lr){7-11} \cmidrule(lr){12-12}
 Finnish (train)              & 1543-1650 & 45897  & 10   & 272  & 1    & 1700-1750     & 47242  & 11   & 214  & 2   & 4289                   \\
Finnish (dev)              & 1543-1650 & 3203  & 12   & 338  & 1    & 1700-1750     & 3351  & 12   & 266  & 2   & 254                   \\
Finnish (test)             & 1543-1650 & 3461  & 12   & 137  & 1    & 1700-1750     & 3264  & 11   & 114  & 2   & 275                   \\
Russian (train)              & 1800-1900 & 1912  & 2  & 12 & 1    & 1950-present     & 4581  & 5   & 19  & 1   & 924                   \\
Russian (dev)              & 1800-1900 & 421  & 2  & 11 & 1    & 1950-present     & 1605  & 8   & 30  & 1   & 201                   \\
Russian (test)              & 1800-1900 & 424  & 2  & 10 & 1    & 1950-present     & 1702  & 8   & 32  & 2   & 211                   \\
German (test)               & 1800–1899 & 584  & 24   & 25  & 20    & 1946–1990 & 568 & 23  & 25 & 14   & 24                   \\
\bottomrule
\end{tabular}
\caption{Statistics of the AXOLOTL-24 datasets. `\#targets' denotes the number of target words; `\#usages' means the total usage count of target words; `avg/u' indicates the average usage count of each target word;  `max/u' indicates the maximum usage count per target word; `min/u' indicates the minimum usage count per target word. }\label{tab:AXOLOTL-24-dataset-statistics}
\end{table*}

The shared task provides datasets for the two subtasks for Finnish, Russian and German languages. These datasets contain dictionary entries such as headwords (target words), the definitions of their meanings, word usages, the positions of the headwords within word usages, and time period (indicating whether word usages belong to an earlier or later time period). 

For Finnish, the dataset is curated from Vanhan kirjasuomen sanakirja (Dictionary of Old Literary Finnish)\footnote{\url{https://kaino.kotus.fi/vks/}} and is split into train, dev and test sets. It includes word usages from earlier and later time periods (before 1700 and after 1700). For Russian, the dataset from an earlier time period is sourced from Explanatory Dictionary of the Living Great Russian Language \citep{dal1955explanatory} while the dataset from a later period is from CODWOE \citep{mickus-etal-2022-semeval}. Again, the dataset is divided into train, dev and test sets. For German, the dataset is collected from DWUG DE Sense \citep{schlechtweg2023human}. The German dataset is only available in the test phase, meaning that no train and dev sets are provided. This setup is to put submitted systems to test in handling an unseen language. We provide data statistics for the AXOLOTL24 shared task in Table \ref{tab:AXOLOTL-24-dataset-statistics}, where the data from earlier and later time periods are treated as two separate corpora.

\paragraph{Implementation details in Subtask 1.} The baseline system is unsupervised, although it still requires a number of hyperparameters. These hyperparameters include a threshold for the minimum similarity between a word usage and a dictionary definition based on their embeddings, as well as parameters required by Affinity Propagation, such as the choice of distance metrics to compute distances between clusters and the number of clustering iterations. The baseline system sets the similarity threshold to 0.3 and keeps the default parameters of Affinity Propagation unchanged for all languages. For our submitted system, two predefined hyperparameters are needed: the similarity threshold as for the baseline system, and the number of nearest neighbors required for generating a semantic graph and computing distances between clusters. After tuning on the development sets, we set the similarity threshold to 0.5 and the number of nearest neighbors to 5 for all languages. On a side note, the baseline system uses the sentence-level encoder LEALLA \citep{mao-nakagawa-2023-lealla} to produce word usage embeddings while our system uses the word-level encoder m-BERT \citep{devlin2018bert} to produce both word and word usage embeddings.

\begin{table*}[h]
\centering 
\footnotesize
\setlength\tabcolsep{5pt} 
\begin{tabular}{lccccccc}
    \toprule
    \multicolumn{1}{l}{} & \multicolumn{1}{l}{} & \multicolumn{2}{c}{Finnish}                & \multicolumn{2}{c}{Russian} & \multicolumn{2}{c}{German}\\
     \cmidrule(lr){3-4}\cmidrule(lr){5-6} \cmidrule(lr){7-8}
    Systems  & \#Entries  & ARI &  macro-F1 & ARI &  macro-F1 & ARI &  macro-F1\\
    \midrule
    deep-change(1)  & 17 & \textbf{0.649} & \textbf{0.760} & \textbf{0.247} & 0.640 &	0.322 &	0.510   \\
    deep-change(2)  & 16 & \textbf{0.649} & \textbf{0.760} & 0.048 & \textbf{0.750} & \textbf{0.521}  &  \textbf{0.740}   \\
    Holotniekat & 4 & 0.596 & 0.630 & 0.043 & 0.660 & 0.298 & 0.610 \\
    ABDN-NLP (Ours) & 2 & 0.553 & 0.590 & 0.009 & 0.570 & 0.102 & 0.300 \\
    \midrule
    Baseline & 5 & 0.023 & 0.230 & 0.079 & 0.260 & 0.022 & 0.130\\
    \bottomrule
\end{tabular}
\caption{Results on the test-phase leaderboard for AXOLOTL-24 Subtask 1. 
}\label{tab:result-track1}
\end{table*}

\begin{table*}[ht]
\centering 
\footnotesize
\setlength\tabcolsep{5pt} 
\begin{tabular}{lccccccc}
    \toprule
    \multicolumn{1}{l}{} & \multicolumn{1}{l}{} & \multicolumn{2}{c}{Finnish}                & \multicolumn{2}{c}{Russian} & \multicolumn{2}{c}{German}\\
     \cmidrule(lr){3-4}\cmidrule(lr){5-6} \cmidrule(lr){7-8}
    Systems    & \#Entries & BLEU &  BERTScore & BLEU &  BERTScore & BLEU &  BERTScore\\
    \midrule
    ABDN-NLP (Ours)  & 3 & \textbf{0.107} & \textbf{0.706} & 0.027 & 0.677 & \underline{0.000} & \textbf{\underline{0.714}}  \\    
    TartuNLP & 1 & 0.028 & 0.679 & \textbf{0.587} & \textbf{0.869} & \textbf{0.010} & 0.630\\
    t-montes & 7 & 0.023 & 0.675 & 0.027 & 0.656 & \textbf{0.010} & 0.650\\
    \midrule
    Baseline  & 6 & 0.033 & 0.403 & 0.005 & 0.377 & 0.000 & 0.490\\
    \bottomrule
\end{tabular}
\caption{Results on the test-phase leaderboard for Subtask 2. Our post-evaluation results are underlined.}\label{tab:result-track2}
\end{table*}

\paragraph{Implementation details in Subtask 2.}
The baseline system is supervised and finetunes the model parameters of XGLM \cite{lin-etal-2022-shot} in the task of generating definitions for word usages. Doing so requires several hyperparameters, including learning rate and weight decay for the Adam optimizer \cite{kingma2014adam}, and the number of epochs for training. The baseline system uses the default parameters of the Adam optimizer and sets the number of epochs to 1.
Our submitted system, on the contrary, is fully unsupervised. For each target word, we take a set of word usages 
identified by our clustering approach in Subtask 1 and prompt LLMs to generate a collective definition for the usages of the target word. A predefined prompt is needed and we provide it in Figure \ref{fig:prompt}. For LLMs, we experiment with GPT-3.5-turbo and GPT-4-turbo, LLaMA-2-7B and LLaMA-3-8B. 

\paragraph{Prompt engineering.} Note that our prompt is created from scratch and refined on a small selection of random instances in the development sets, meaning that our prompt is not optimal for the entire sets in any language. Our refinement process starts with an English prompt to instruct LLMs to generate Finnish, Russian and German definitions; however, LLMs often generate English definitions for non-English word usages; we address this by translating the English prompt into Finnish, Russian and German via Google Translate. Other factors for refinement include (a) the length of a definition, (b) determining when to stop generation in order to ensure that generated definitions are comparable in length to the ground-truth counterparts, and (c) the number of word usages for LLMs to generate a collective definition.

\paragraph{Evaluation.} For Subtask 1, 
the Adjusted Rand Index (ARI) \citep{hubert1985comparing} and macro-F1 score are the two evaluation metrics for reporting and comparing system performances. 
ARI calculates how much a pair of word usages from the predictions belong to the same sense ID (or different sense IDs) as they should, while the macro-F1 score computes the precision and recall of word usages for each sense ID and then averages these scores across all sense IDs.
Note that F1 only considers old senses in the ``new'' time period, meaning that mappings of word usages to novel senses are not evaluated. ARI considers both novel and old senses in the ``new'' time period.

For Subtask 2, generated definitions for those usages with novel senses are compared to their ground-truth counterparts by computing similarities between definition pairs.
The AXOLOTL-24 shared task uses both lexical-based and embedding-based metrics to compute definition pair similarities. The metrics considered are BLEU \citep{papineni2002bleu} and BERTScore \citep{Zhang2020BERTScore:}. Other metrics appropriate for doing so include MoverScore \citep{zhao-etal-2019-moverscore}, BlonDe \citep{jiang-etal-2022-blonde} and DiscoScore \citep{zhao-etal-2023-discoscore}. The latter two metrics have shown to be well-suited for computing long-text pair similarities, particularly useful when dealing with lengthy definitions.

\section{Results}
We present the results of our systems and analyses on LLMs. Case studies are shown in Appendix \ref{sec:appendix}.

\paragraph{Subtask 1.} We made two submissions for Subtask 1, with minor difference between them. The only difference is that the second submission includes additional predictions for the unseen German language. 
Table \ref{tab:result-track1} compares the results of our system and other teams. We see that our system, based on the unsupervised graph-based clustering approach, outperforms the unsupervised baseline system by a large margin in all the languages. 
We observe a big performance drop for the German language compared to other two languages. One of the reasons for this is due to historical data issues. Unlike the Russian and Finnish corpus usages---which have been carefully preprocessed by AXOLOTL-24 organizers, German usages are not cleaned up and contain spelling variations (e.g., \textit{n\"othig} instead of \textit{n\"otig}), OCR errors, escaping double quotes and others. These issues would incur out-of-vocabulary tokens, potentially resulting in poor performance.

Lastly, our system performs poorly in terms of ARI on both Russian and German test sets, despite having better scores in macro-F1. 
The performance gap between F1 and ARI attributes to the scope mismatch between the two metrics: new sense IDs are excluded when computing F1, whereas both old and novel sense IDs are considered when computing ARI. This means unlike ARI, F1 would not penalize wrong prediction of novel sense IDs. As a result, although our system performs poorly for novel sense predictions in Russian (see the ARI\_new result in Table \ref{tab:ari}), the F1 result (F1$=$0.570) is still quite high.

\begin{table}[h]
\centering 
\footnotesize
\setlength\tabcolsep{5pt} 
\begin{tabular}{lccccccc}
    \toprule
    
    Metrics  & Finnish &  Russian & German \\
    \toprule
    macro-F1 & 0.590 & 0.570 & 0.300 \\
    \midrule
    ARI & 0.596 & 0.043 & 0.298 \\
    ARI\_new & 0.633 & 0.039 & 0.524\\
    ARI\_old & 0.619 & 0.754 & 0.260\\
    \bottomrule
\end{tabular}
\caption{Post-evaluation results of our system on the test-phase leaderboard for AXOLOTL-24 Subtask 1. 
ARI\_new considers new sense IDs only, while ARI\_old focuses on old sense IDs.
}\label{tab:ari}
\end{table}

Note that the results from our system and other teams are not directly comparable as the system details of other teams are missing. For instance, it remains unclear whether their systems are unsupervised or not. Overall, we see the deep-change system achieves the best performance in all the three languages (including the unseen German language where train and dev sets are unavailable); however, their achievement is made through a total of 33 submissions and the leaderboard only reports their best performance; this indicates overfitting. 

\begin{table}[h]
\centering 
\footnotesize
\setlength\tabcolsep{3pt} 
\begin{tabular}{lcccc}
    \toprule
    \multicolumn{1}{l}{} & \multicolumn{2}{c}{Finnish}                & \multicolumn{2}{c}{Russian} \\
     \cmidrule(lr){2-3}\cmidrule(lr){4-5} 
    LLMs    & BLEU &  BERTScore & BLEU &  BERTScore\\
    \midrule
    Baseline &  \textbf{0.248} & 0.607  & \textbf{0.886}  & 0.595 \\
    \midrule
    GPT-3.5-turbo & 0.022 & 0.640  & 0.035 &  0.676 \\
    GPT-4-turbo   & 0.025 & \textbf{0.658}  & 0.036 &  \textbf{0.678} \\
    LLaMA-2-7B & 0.013 & 0.611  & 0.024 & 0.604\\
    LLaMA-3-8B & 0.013 &  0.603 & 0.021 & 0.601\\
    \bottomrule
\end{tabular}
\caption{Comparing LLMs on the dev set in Subtask 2.}\label{tab:llms}
\end{table}

\paragraph{Subtask 2.} We refined our prompts for instructing GPT-3.5-turbo. 
This results in three submissions we made for Subtask 2, where the prompts in our final submission yield the best performance on the randomly selected instances from the Finnish and Russian dev sets. Note that the final prompts are the Finnish, Russian and German translations from the English version (see Figure \ref{fig:prompt}).

Despite not using train sets, our unsupervised system, based on GPT-3.5-turbo, considerably outperforms the supervised baseline system in all setups (see Table \ref{tab:result-track2}). This might be because the train sets are not large enough for fine-tuning XGLM \cite{lin-etal-2022-shot}. When compared with other teams, our system ranks first for Finnish and German, and ranks second for Russian. Again, it is unclear whether other teams take advantage of the train sets, and thus the direct comparison with other systems is not meaningful.

\paragraph{Comparison of LLMs.} Figure \ref{tab:llms} compares the results of several LLMs. Overall, we observe that our unsupervised system based on LLMs greatly outperforms the supervised baseline system in terms of BERTScore. However, our system performs worse than the baseline in BLEU. This is because our generated definitions are not lexically but semantically similar to their ground-truth counterparts. The reason for this is the following:
BLEU cannot recognize text pair similarity when there is no lexical overlap between them \citep{10.1162/coli_a_00322}. This is particularly problematic when dealing with morphologically rich languages like Russian and Finnish. In such languages, high-quality generated definitions might differ greatly from ground-truth definitions in morphological forms; in this case, BLEU would wrongly assign low scores to high-quality definitions due to the absence of lexical overlap. This is demonstrated by our results, where BLEU scores (0.02-0.03) mean very few lexical overlaps between the generated and ground-truth definitions while BERTScore (0.65-0.67) suggest that definition pairs are indeed semantically similar. 

Additionally, we observe the supervised baseline system performs best in terms of BLEU, particularly for Russian. This means the generated definitions are lexically similar to the ground-truth. This might be attributed to
the memorization of training sets. We see that many ground-truth definitions contain words from corpus usages. During training, the baseline system might have learned to prioritize the use of words from corpus usages when generating definitions. Lastly, although GPT-4-turbo has shown to greatly outperform GPT-3.5-turbo in many NLP tasks, we demonstrate that the superiority of GPT-4-turbo is not considerable in Subtask 2, especially for Russian, so is the case for LLaMA-2-7B and LLaMA-3-8B.

\section{Limitations}

\paragraph{Dataset size.} The datasets provided in the shared task are quite small and contain very few word usages for each headword on average. This is indeed expected as the datasets are sourced from hand-crafted dictionaries where lexicographers only collect a small number of word usages for each dictionary sense entry due to the costly mapping process. Here we argue that it would be better to use such datasets only for evaluation purposes, rather than for dividing them into train sets. Furthermore, we call for an additional database containing a large amount of word usages for each headword to support the development of unsupervised systems, as we see their potential demonstrated by our unsupervised system, which greatly outperformed the supervised baseline system in Subtask 2. 

\paragraph{Text encoder.} Our system relies on m-BERT \citep{devlin2018bert}, a text encoder invented five years ago, to produce embeddings for both word usages and words in Subtask 1. In recent years, many text encoders \citep{ni-etal-2022-sentence, neelakantan2022text} have been introduced and shown to perform much better than m-BERT in various NLP tasks. Other encoders such as XL-LEXEME \citep{cassotti-etal-2023-xl} specialized in capturing lexical semantic changes also meet our needs.

\paragraph{Data contamination.} The works by \citet{balloccu-etal-2024-leak, ravaut2024much} show that the results of LLMs can be misleading due to the data contamination issue, i.e., that test sets are included in the training data of LLMs. This issue might be present in the AXOLOTL-24 test sets for the two reasons: (a) the source base of the test sets is publicly accessible and (b) LLMs do not document their training data at all. Thus, it is unclear whether the headwords, word usages, and definitions in the test sets have been exposed to LLMs. Future work should design a measure to calculate data contamination rates of LLMs on the AXOLOTL-24 datasets. 

\section{Conclusions}
In this work, we presented our system that automates the process of identifying novel word meanings not covered in dictionaries and generating their definitions. We  evaluated our system in the AXOLOTL-24 shared task.
Our results show that supervision is not always useful: Without access to train sets, our unsupervised system still greatly outperforms the supervised baseline system, as well as other team submissions in Subtask 2---which demonstrates the potential of LLMs in generating definitions for novel word usages; however, the uncertainty as to whether the AXOLOTL-24 test sets are included in  the training data for pre-training LLMs calls for careful investigation in the future.

\section*{Acknowledgements}
We thank the anonymous reviewers for their thoughtful feedback that greatly improved the texts. Dominik Schlechtweg has been funded by the research program `Change is Key!' supported by Riksbankens Jubileumsfond (under reference number M21-0021).

\bibliography{custom}

\appendix
\clearpage
\section{Appendix}
\label{sec:appendix}

\paragraph{Case studies.} Figures \ref{fig:case-study-1} and \ref{fig:case-study-2} compare generated and ground-truth definitions for the two target words sampled from the Russian dev set. For the first target word, the generated definition by GPT-3.5-turbo is quite similar to the ground-truth definition. We suspect that the word `radioactive' in the corpus usage suggests that the location is likely to be a burial ground. We test this hypothesis by removing the word `radioactive' and prompting GPT-3.5-turbo again: the generation definition then becomes ``a burial ground or cemetery'' (English translation)---which is too general and refers to a non-metaphorical scenario where people are buried underground, whereas ``radioactive burial ground'' could mean metaphorically a site for disposing of radioactive waste.

Consider the second word, which is computer slang meaning ``to make something inaccessible''. Interestingly, GPT-3.5-turbo did not provide any guess on the definition of the word usage, and just acknowledged that this is a Russian word without giving further details. This could be because GPT-3.5-turbo lacks knowledge of the cybersecurity term `DDoS' (it means a denial-of-service attack), and thus it did not provide any guess for the definition. This analysis, however, is only based on two cases. Future work could include a due diligence investigation of wrongly generated definitions, such as categorizing incorrect definitions.

\begin{figure}[h]
    \begin{minipage}{\textwidth}
    \begin{verbatim}
#A usage for the word: [могильник]

Якобы тут, возле Черниховки, находится 
радиоактивный могильник.

(English Translation): Allegedly here, near Chernik-
hovka, there is radioactive burial ground.

[Generated Definition by GPT-3.5-turbo]:
Могильник - место захоронения радиоактивных 
отходов или погибших. 

(English Translation): Burial ground - a burial place 
for radioactive waste or dead.

[Ground-truth Definition]:
спец. место захоронения радиоактивных отходов; 
специальное сооружение для такого захоронения.

(English Translation): Special burial site for radio-
active waste; special structure for such burial.

[Evaluation]: BLEU: 21.2      BERTScore: 0.79
    \end{verbatim}    
    \end{minipage}
    \caption{A well-generated definition in Russian.}
    \label{fig:case-study-1}
\end{figure}

\begin{figure}
    \begin{minipage}{\textwidth}
    \begin{verbatim}
# A usage for the headword: [положить]

Также впустую ушли две недели ддоса главной 
страницы пикаперов, хотя форум положить удалось.

(English Translation): Also, two weeks of DDoSing the 
main page were wasted pick-up artists' pages, although 
they managed to put down the forum.

[Generated Definition by GPT-3.5-turbo]:
язык, используемый в России и других странах.

(English Translation): A language used in Russia and 
other countries.

[Ground-truth Definition]: 
комп. жарг. привести в нерабочее состояние, 
«уронить», сделать недоступным что-либо.

(English Translation): A computer slang referring to 
something inoperative and inaccessible.

[Evaluation]: BLEU: 3.38     BERTScore: 0.59
    \end{verbatim}    
    \end{minipage}
    \caption{A poorly-generated definition in Russian.}
    \label{fig:case-study-2}
\end{figure}

\paragraph{Our prompt.} Figure \ref{fig:prompt} illustrates the prompt used to instruct GPT-3.5-turbo to generate definitions in English. 
\begin{figure}[h]   
    \begin{minipage}{\textwidth}
    \begin{verbatim}
[Instruction]:
Imagine that you are a lexicographer, given a headword 
{target_word} in {lang}, write the dictionary definition 
of its usage in the following quotations:

1. First quotation
2. Second quotation

[Requirements]:
The definition you create should be brief. A maximum 
of ten words is allowed. The definition ends at the 
first period.

[Response]:
Definition (string): {definition}
    \end{verbatim}    
    \end{minipage}
    \caption{An illustration of our prompt used to instruct GPT-3.5-turbo to generate dictionary-like definitions, where `quotation' is synonymous of `word usage'.}
    \label{fig:prompt}
\end{figure}

\end{document}